\documentclass[a4paper,twoside]{article}

\usepackage{epsfig}
\usepackage{subcaption}
\usepackage{calc}
\usepackage{amssymb}
\usepackage{amstext}
\usepackage{amsmath}
\usepackage{siunitx}
\usepackage{amsthm}
\usepackage{amssymb}
\usepackage{multicol}
\usepackage{pslatex}
\usepackage{apalike}
\usepackage{caption}
\usepackage{subcaption}
\usepackage{ragged2e}
\usepackage{graphics}
\usepackage{soul}
\usepackage{textcomp}
\usepackage[bookmarks=false]{hyperref}
\newlength\shlength

\DeclareMathOperator*{\argmin}{arg\,min}
\DeclareMathOperator*{\argmax}{arg\,max}

\usepackage{array}

\usepackage{color, colortbl}
\definecolor{LightCyan}{rgb}{0.88,1,1}
\usepackage[first=0,last=9]{lcg}

\usepackage{makecell}
\usepackage{array}
\newcolumntype{P}[1]{>{\centering\arraybackslash}p{#1}}

\usepackage{comment}
\usepackage{xcolor}

\usepackage{SCITEPRESS} % Please add other packages that you may need BEFORE the SCITEPRESS.sty package.

\begin{document}

\title{Sim-to-Real Transfer with Incremental Environment Complexity \\ for Reinforcement Learning of Depth-Based Robot Navigation}

\author{\authorname{Thomas Chaffre\sup{1,2}, Julien Moras\sup{3}, Adrien Chan-Hon-Tong\sup{3} and Julien Marzat\sup{3}}
\affiliation{\sup{1}Lab-STICC UMR CNRS 6285, ENSTA Bretagne, Brest, France}
\affiliation{\sup{2}School of Computer Science, Engineering and Mathematics, Flinders University, Adelaide, SA, Australia}
\affiliation{\sup{3}DTIS, ONERA - The French Aerospace Lab, Universit\'e Paris Saclay, F-91123 Palaiseau, France}
\email{julien.moras@onera.fr, adrien.chan\_hon\_tong@onera.fr, julien.marzat@onera.fr}
}


\abstract{Transferring learning-based models to the real world remains one of the hardest problems in model-free control theory. Due to the cost of data collection on a real robot and the limited sample efficiency of Deep Reinforcement Learning algorithms, models are usually trained in a simulator which theoretically provides an infinite amount of data. Despite offering unbounded trial and error runs, the reality gap between simulation and the physical world brings little guarantee about the policy behavior in real operation. Depending on the problem, expensive real fine-tuning and/or a complex domain randomization strategy may be required to produce a relevant policy. 
In this paper, a Soft-Actor Critic (SAC) training strategy using incremental environment complexity is proposed to drastically reduce the need for additional training in the real world. The application addressed is depth-based mapless navigation, where a mobile robot should reach a given waypoint in a cluttered environment with no prior mapping information. 
Experimental results in simulated and real environments are presented to assess quantitatively the efficiency of the proposed approach, which demonstrated a success rate twice higher than a naive strategy.}

\onecolumn \maketitle \normalsize \setcounter{footnote}{0} \vfill

\section{\uppercase{Introduction}} \label{sec:introduction}

\noindent State-of-the-art algorithms are nowadays able to provide solutions to most elementary robotic problems like exploration, mapless navigation or Simultaneous Localization And Mapping (SLAM), under reasonable assumptions~\cite{Cadena2016PastPA}. However, robotic pipelines are usually an assembly of several modules, each one dealing with an elementary function (e.g. control, planning, localization, mapping) dedicated to one technical aspect of the task. Each of these modules usually requires expert knowledge to be integrated, calibrated, and tuned. Combining several elementary functions into a single grey box module is a challenge but is an extremely interesting alternative in order to reduce calibration needs or expertise dependency. Some of the elementary functions can raise issues in hard cases (e.g., computer vision in weakly textured environment or varying illumination conditions). Splitting the system into a nearly optimal control module processing a coarse computer vision mapping output may result in a poorer pipeline than directly using a map-and-command module which could achieve a better performance trade-off. \\
\noindent For this reason, there is a large academic effort to try to combine several robotic functions into learning-based modules, in particular using a deep reinforcement strategy as in~\cite{zamora2016extending}. A limitation of this approach is that the resulting module is task-dependent, thus usually not reusable for other purposes even if this could be moderated by multi-task learning. A more serious limit is that learning such a function requires a large amount of trial and error. Training entirely with real robots is consequently unrealistic in practice considering the required time (even omitting physical safety of the platform during such learning process where starting behavior is almost random). On the other hand, due to the reality gap between the simulator and the real world, a policy trained exclusively in simulation is most likely to fail in real conditions~\cite{DulacArnold2019ChallengesOR}. Hence, depending on the problem, expensive real fine-tuning, and/or a complex domain randomization strategy may be required to produce a relevant policy. The explainability and evaluation of safety guarantees provided by such learning approaches compared to conventional methods also remain challenging issues~\cite{juozapaitis2019explainable}.

\begin{figure}[b!]
\vspace{-0.25cm}
\captionsetup{justification=justified}
\centering
	\begin{subfigure}{0.55\columnwidth}
	\captionsetup{justification=justified}
		\centering
        \includegraphics[height=4.2cm,width=0.99\linewidth]{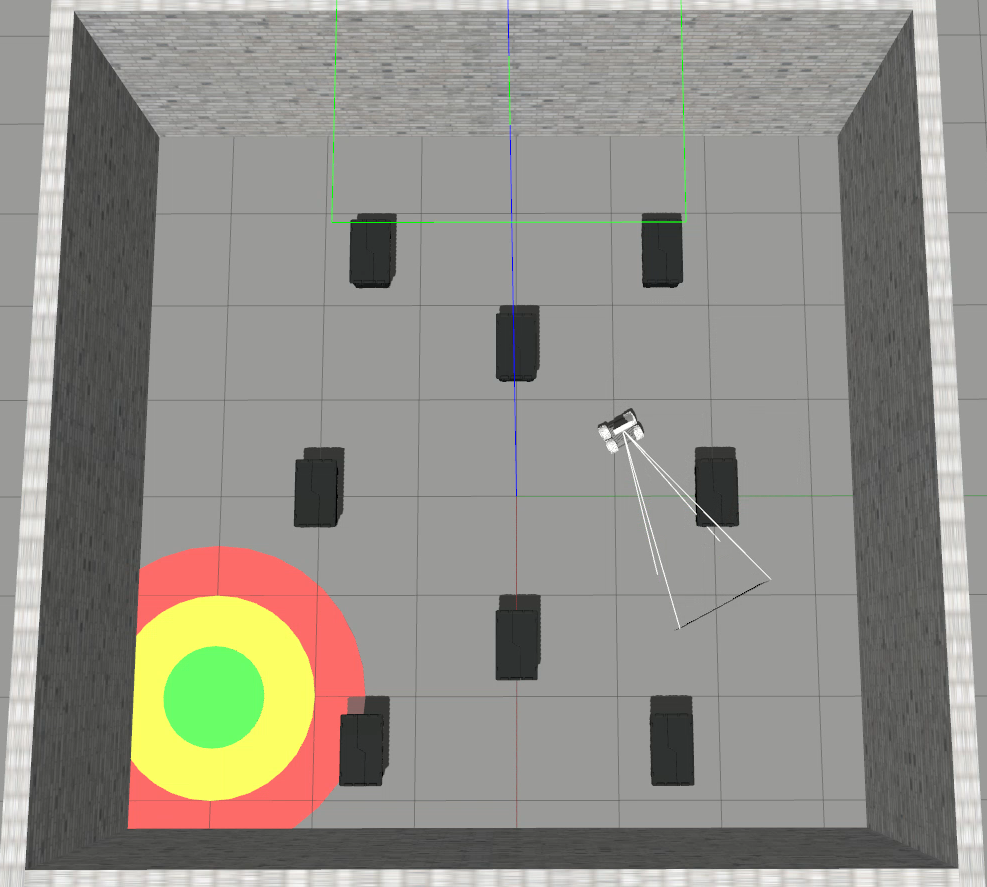}
        \caption{Robot learning in the ROS-Gazebo simulated environment.}
    \end{subfigure}\hspace{0.125cm}
    \begin{subfigure}{0.42\columnwidth}
    \vspace{0.15cm}
    \captionsetup{justification=justified}
        \includegraphics[height=4.2cm, width=0.99\linewidth]{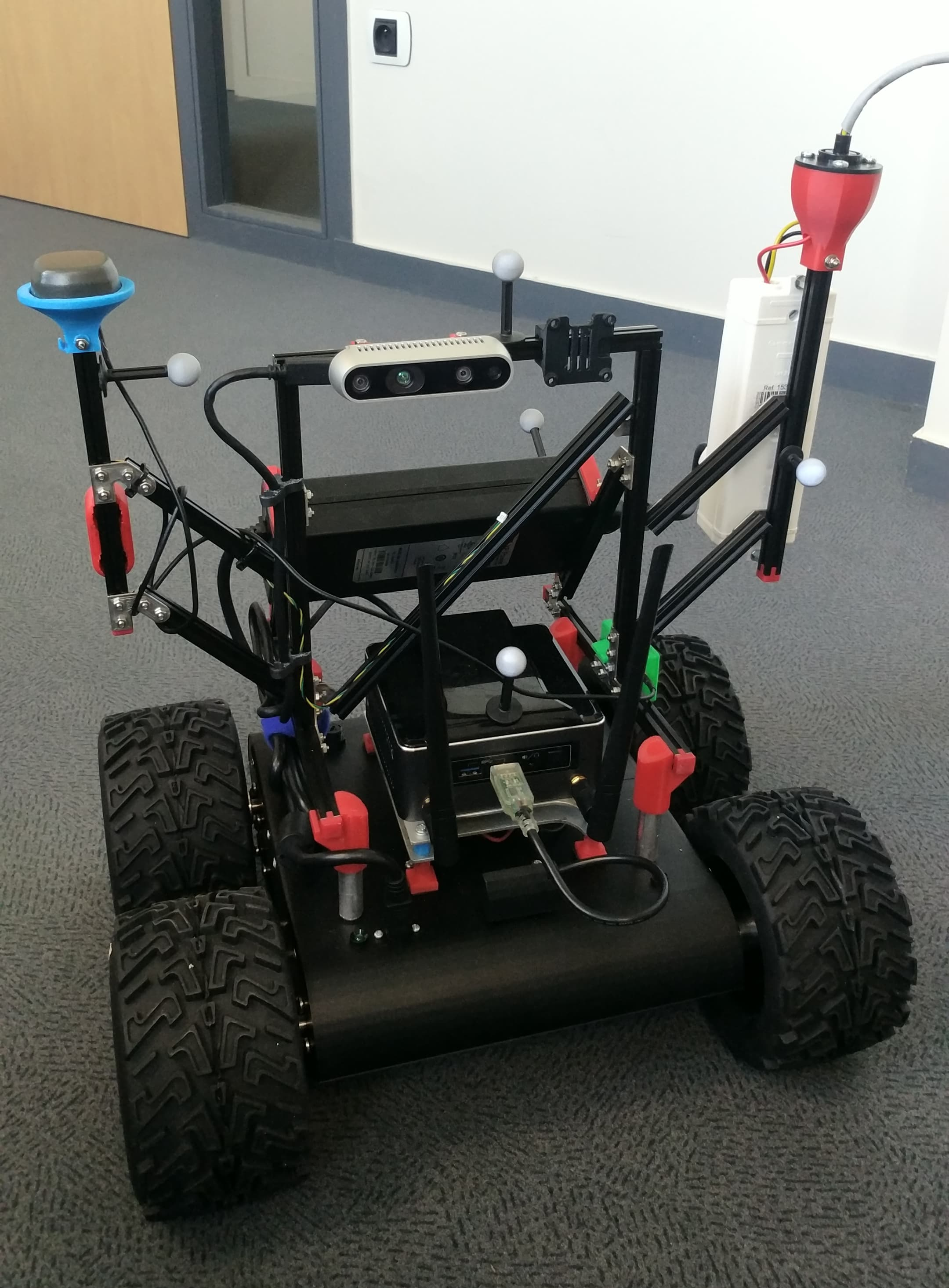}
        \caption{Cable-powered Wifibot with depth sensor.}
    \vspace{0.15cm}
    \end{subfigure}
\begin{subfigure}{\columnwidth}
    \captionsetup{justification=justified}
    \centering
    \includegraphics[width=0.99\columnwidth]{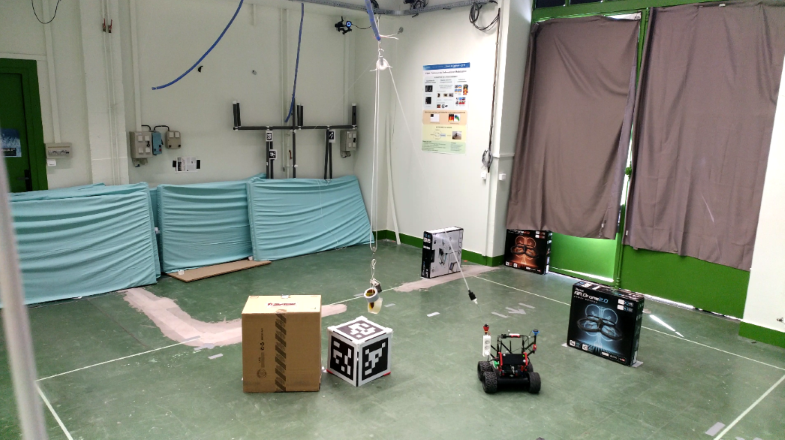}
    \caption{Wifibot navigating in the real environment.}
    \end{subfigure}
    \vspace{0.1cm}
    \caption{Illustration of simulated and real environments for reinforcement learning of depth-based robot navigation.\label{fig:wifibot}}
    %\vspace{-0.5cm}
\end{figure}

\noindent In this work, we address the problem of robot navigation in an uncharted cluttered area with Deep Reinforcement Learning. In this context we consider as a benchmark the task where a robotic agent (here a Wifibot Lab v4 mobile robot, see Figure~\ref{fig:wifibot}) has to reach a given target position in a cluttered room. 
At the beginning of each episode, the robot starts at the same location but with a random orientation and gets the target position coordinates. The robot is equipped with a perception sensor providing a dense depth map (here an Intel RealSense D435), it has access to its current position in the environment and has to control the speed of its wheels (which are the continuous outputs of the proposed learning algorithm). The proposed training method is based on the Soft Actor-Critic method~\cite{haarnoja2018soft} coupled with incremental environment complexity. The latter refers to a technique which consists in splitting the desired mission requirements into several environments, each one representing a different degree of complexity of the global mission. Furthermore, an experimental setup is also proposed in this paper for testing or learning continuation in the real environment without human supervision.

\noindent The paper is organized as follows. Section \ref{sec:related_work} presents related work in robotic navigation and reinforcement learning. Section \ref{sec:method} details the proposed method, particularly the agent structure, the reward generation and the learning procedure. Finally, experiments and results in both simulated and real environments are presented in Section \ref{sec:expe_result}.

\section{\uppercase{Related Work}} \label{sec:related_work}
\noindent Classical methods for robot autonomous navigation are usually based on a set of several model-based blocks that perform SLAM~\cite{mur2017orb,engel2015large,wurm2010octomap} and waypoint navigation algorithms~\cite{sucan2012the-open-motion-planning-library,kamel2017model,bak2001path}. The latter use either reactive methods, which are limited in horizon and sometimes trapped in local minima, or a combination of a trajectory planner using an additional mapping algorithm and a tracking controller. Trajectory planning is usually highly time-consuming and has difficulties to adapt to real-time computational constraints. It could be hinted that learning-based strategies will be able to achieve an implicit computational and informational trade-off between these techniques. In~\cite{Mishkin.2019}, classic and learning-based navigation systems have been compared. The modular navigation pipeline they proposed divides the navigation process into 4 sub-tasks: mapping, localization, planning and locomotion. They demonstrated that the classical system outperforms the learned agent when taking as input RBG-D values. In contrast, the classical method is very sensitive to the global set of modalities (i.e. without depth information, it fails catastrophically). Although model-based\footnote{Here the term model does not correspond to the one used in Section \ref{sec:method} but refers to whether or not a dynamical model of the controlled system is used in the control strategy whereas in RL, it is the model of the environment that would be provided to the RL algorithm.} methods usually work well in practice, they demand expert knowledge and do not allow to tackle missions requiring great interaction with the environment. On the other hand, model-free approaches (in particular reinforcement learning) have shown impressive progress in gaming~\cite{silver2017mastering}, and are beginning to be widely applied for robotic tasks involving computer vision (see \cite{carrio2017review} for a review). This paradigm has also been successfully applied to video game environments~\cite{mnih2013playing,kempka2016vizdoom,lample2017playing}, where the input space is similar to the control space of a robot. However these results cannot be readily applied in robotics, because these strategies have been learned and tested in the same virtual environment and moreover the reward (game score) is explicit.\\
\noindent Dedicated Deep Reinforcement Learning strategies have already been applied to different robotic platforms and functions. In~\cite{chiang2019learning}, point-to-point and path-following navigation behaviors that avoid moving obstacles were learned using a deep reinforcement learning approach, with validation in a real office environment. In~\cite{xie2017towards}, a dueling architecture based on a deep double-Q network (D3QN) was proposed for obstacle avoidance, using only monocular RGB vision as input signal. A convolutional neural network was constructed to predict depth from raw RGB images, followed by a Deep Q-Network consisting of a convolutional network and a dueling network to predict the Q-Value of angular and linear actions in parallel. They demonstrated the feasibility of transferring visual knowledge from virtual to real and the high performance of obstacle avoidance using monocular vision only.  In~\cite{zamora2016extending}, a robot patrolling task was successfully learned in simulation. However, this strategy puts the emphasis on not hitting any obstacle more than anything else, therefore the system is not strongly forced to \textit{take risks} (which is required when heading to a designated destination). Visual object recovery \cite{sampedro2019fully} has also been considered, the task being defined as reaching an object described by an appearance where all objectives are described in a common vocabulary: proximity to obstacle and proximity of target are embedded in the visual space directly. These two approaches have only been validated in a simulation environment, therefore an undetermined amount of work remains to transfer them to a real robot and environment. In~\cite{ecmr2019deepnav}, a navigation task to an image-defined target was achieved using an advantage actor-critic strategy, and a learning process with increased environment complexity was described. The method we propose is similar in spirit to this one, but our contribution addresses mapless navigation as in~\cite{Tai2017}, which forces the system to take more risks: the system fails if it does not reach the target sufficiently fast, so going closer to obstacles (without hitting them) should be considered. Also, in this task, the system has to process both metric and visual information: distance to obstacles should be perceived from sensor measurements (image, depth), while the target is given as a coordinate. We propose a new learning strategy to tackle this problem, similar to Curriculum Learning (CL) \cite{Elman1993LearningAD,Bengio2009CurriculumL} but easier to implement. CL aims at training a neural network more efficiently by using the concept of curriculum, a powerful tool inspired by how humans progressively learn from simple concepts to harder problems. Recent studies \cite{Zaremba2014LearningTE,Rusu2017SimtoRealRL,OpenAI2019SolvingRC} on the application of this method in the robotic field have shown promising outcomes. 
A drawback of these approaches is the need for heavy simulation, however there is little alternative: fine-tuning in real life seems to be a candidate, but as the fine-tuning database may be quite limited, it is hard (with a deep model) to avoid overfitting and/or catastrophic forgetting. In this paper, we study the behavior of the policy transferred from a simulated to a real environment, with a dedicated hardware setup for unsupervised real testing with a mobile robot. It turns out that depth-based mapless navigation does not seem to require a too heavy domain randomization or fine-tuning procedure with the proposed framework based on incremental complexity. 

\section{\uppercase{sac-based navigation framework} \label{sec:method}}

\subsection{Preliminaries}\label{SAC}

For completeness, we recall here the \textit{Policy Gradient} \cite{Sutton1999PolicyGM} point of view in which we aim at modeling and optimizing the policy directly. More formally, a policy function $\pi$ is defined as follows: 
$$\pi_\theta:S \rightarrow A$$
Where $\theta$ is a vector of parameters, $S$ is the state space and $A$ is the action space.
The vector $\theta$ is optimized and thus modified in training mode, while it is fixed in testing mode.
The performance of the learned behavior is commonly measured in terms of success rate (number of successful runs over total number of runs). Typically for mapless navigation, a successful run happens if the robot reaches the targeted point without hitting any obstacle in some allowed duration. In training mode, the objective is to optimize $\theta$ such that the success rate during testing is high. However, trying to directly optimize $\theta$ with respect to the testing success rate is usually sample inefficient (it could be achieved using e.g. CMA-ES~\cite{salimans2017evolution}). Thus, the problem is instead modelled as a Markov process with state transitions associated to a reward. The objective of the training is therefore to maximize the expected (discounted) total reward:
\begin{equation*}
\underset{\theta}{\min} \underset{t\in {1,...,T}}{\sum} \left( \left( \underset{\tau\in {t,...,T}}{\sum} r_{\tau}\gamma^{\tau-t} \right) \log\left(\pi_{\theta}(a_t|s_t) \right)\right)    
\end{equation*}\\
\noindent Direct maximization of the expected reward is a possible approach, another one consists in estimating the expected reward for each state, and they can be combined to improve performance. A turning point in the expansion of RL algorithms to continuous action spaces appeared in \cite{Lillicrap2016ddpg} where Deep Deterministic Policy Gradient (\textbf{DDPG}) was introduced, an actor-critic model-free algorithm that expanded Deep Q-Learning to the continuous domain. This approach has then be improved in~\cite{haarnoja2018soft} where the Soft Actor Critic (\textbf{SAC}) algorithm was proposed: it corresponds to an actor-critic strategy which adds a measure of the policy entropy into the reward to encourage exploration. Therefore, the policy consists in maximizing simultaneously the expected return and the entropy:
\begin{equation}
J(\theta)=\sum_{t=1}^T\mathbb{E}_{(s_t,a_t)\sim \rho_{\pi_\theta}}[r(s_t,a_t)+\alpha H(\pi_\theta(.|s_t))]
\end{equation}where

\begin{equation}
H(\pi_\theta(.|s))=-\sum\limits_{a\in A}\pi_\theta(a)\log \pi_\theta(a|s)
\end{equation}
\noindent The term $H(\pi_\theta)$ is the entropy measure of policy $\pi_\theta$ and $\alpha$ is a temperature parameter that determines the relative importance of the entropy term. Entropy maximization leads to policies that have better exploration capabilities, with an equal probability to select near-optimal strategies. SAC aims to learn three functions, a policy $\pi_\theta(a_t|s_t)$, a soft Q-value function $Q_w(s_t,a_t)$ parameterized by $w$ and a soft state-value function $V_\Psi(s_t)$ parameterized by $\Psi$. The Q-value and soft state-value functions are defined as follows:
\begin{equation} \label{gamma}
Q_{w}(s_t,a_t)=r(s_t,a_t)+\gamma\mathbb{E}_{s_{t+1}\sim\rho_\pi(s)}[V_{\Psi}(s_{t+1})]
\end{equation}
\begin{equation}
V_{\Psi}(s_t)=\mathbb{E}_{a_t\sim\pi}[Q_{w}(s_t,a_t)-\alpha\log{\pi_{\theta}(a_t|s_t)}]
\end{equation}\\
Theoretically, we can derive $V_\Psi$ by knowing $Q_w$ and $\pi_\theta$ but in practice, trying to also estimate the state-value function helps stabilizing the training process. The terms $\rho_\pi(s)$ and $\rho_\pi(s,a)$ denote the state and the state-action marginals of the state distribution induced by the policy $\pi(a|s)$. The Q-value function is trained to minimize the soft Bellman residual:
\begin{equation}
\begin{split}
&J_Q(w)=\mathbb{E}_{(s_t,a_t)\sim R}[\frac{1}{2}(Q_w(s_t,a_t)-(r(s_t,a_t) \\ & +\gamma\mathbb{E}_{s_{t+1}\rho_\pi(s)}[V_{\hat{\Psi}}(s_{t+1})]))^2]
\end{split}
\end{equation}\\
The state-value function is trained to minimize the mean squared error:
\begin{equation}
\begin{split}
&J_V(\Psi)=\mathbb{E}_{s_t\sim R}[\frac{1}{2}(V_\Psi(s_t)-\mathbb{E}[Q_w(s_t,a_t)  \\
&-\log{\pi_\theta}(a_t,s_t)])^2]
\end{split}
\end{equation}
The policy is updated to minimize the Kullback-Leibler divergence:
\begin{equation}
\begin{split}
&\pi_{new}=\argmin\limits_{\pi'\in \prod}D_{KL}(\pi'(.|s_t),\exp(Q^{\pi_{old}}(s_t,.)\\
&-\log Z^{\pi_{old}}(s_t)))
\end{split}
\end{equation}
We use the partition function $Z^{\pi_{old}}(s_t)$ to normalize the distribution and while it is intractable in general, it does not contribute to the gradient with respect to the new policy and can thus be neglected. This update guarantees that $Q^{\pi_{new}}(s_t,a_t)\geq Q^{\pi_{old}}(s_t,a_t) $, the proof of this lemma can be found in the Appendix B.2 of \cite{haarnoja2018soft}. 

\noindent Despite performing well in simulation, the transfer of the obtained policy to a real platform is often problematic due to the reality gap between the simulator and the physical world (which is triggered by an inconsistency between physical parameters and incorrect physical modeling).
Recently proposed approaches have tried to either strengthen the mathematical model (simulator) or increase the generalization capacities of the model \cite{ruiz2019,kar2019}. Among the existing techniques that facilitate model transfer, domain randomization (\textbf{DR}) is an unsupervised approach which requires little or no real data. It aims at training a policy across many virtual environments, as diverse as possible. By monitoring a set of $N$ environment parameters with a configuration $\Sigma$ (sampled from a randomization space, $\Sigma\in\Xi\in\mathbb{R}^N$), the policy $\pi_\theta$ can then use episode samples collected among a variety of configurations and as a result learn to better generalize. The policy parameter $\theta$ is trained to maximize the expected reward $R$ (of a finite trajectory) averaged across a distribution of configurations:
\begin{equation} \label{eq:dr}
    \theta^*=\argmax\limits_{\theta}\mathbb{E}_{\Sigma\sim\Xi}\left[\mathbb{E}_{\pi_\theta,\tau\sim e_\Sigma}[R(\tau)]\right]
\end{equation}
\noindent where $\tau$ is a trajectory collected in the environment randomized by the configuration $\Sigma$. In~\cite{vuong2019pick}, domain randomization has been coupled with a simple iterative gradient-free stochastic optimization method (Cross Entropy) to solve~\eqref{eq:dr}. Assuming the randomization configuration is sampled from a distribution parameterized by $\phi$, $\Sigma\sim P_\phi(\Sigma)$, the optimization process consists in learning a distribution on which a policy can achieve maximal performance in the real environment $e_{real}$:
\begin{equation}
\hspace{10px}\phi^*=\argmin\limits_{\phi} \cal{L} (\pi_{\theta^*(\phi)};\text{$e_{real}$}),
\end{equation}where
\begin{equation}
\theta^*(\phi)=\argmin\limits_{\phi}\mathbb{E}_{\Sigma\sim P_\phi(\Sigma)}[\cal{L}(\pi_\theta;\text{$e_{\Sigma}$)}]
\end{equation}
\newpage\noindent The term $\cal{L(\pi\text{,e})}$ refers to the loss function of policy $\pi$ evaluated in environment $e$. Since the ranges for the parameters are hand-picked in this uniform DR, it can be seen as a manual optimization process to tune $\phi$ for the optimal $\cal{L}(\pi_\theta\text{;$e_{real}$})$. 
The effectiveness of DR lies in the choice of the randomization parameters. In its original version \cite{Sadeghi2016,Tobin2017}, each randomization parameter $\Phi_i$ was restricted to an interval ${\Phi_i \in [\Phi_i^{low};\Phi_i^{high}], i=1,\dots,N}$. The randomization parameters can control appearance or dynamics of the training environment.
\subsection{Proposed Learning Architecture} \label{sec:RLset}

\subsubsection{State and observation vectors} \label{obsvectors}
The problem considered is to learn a policy to drive a mobile robot (with linear and angular velocities as continuous outputs) in a cluttered environment, using the knowledge of its current position and destination (external inputs) and the measurements acquired by its embedded depth sensor.
The considered state is defined as:
\begin{equation}
s_t=(o_t,p_t,h_t,a_{t-1})
\end{equation}\\
\noindent where $o_t$ is the observation of the environment from the depth sensor, $p_t$ and $h_t$ are respectively the relative position and heading of the robot toward the target, $a_{t-1}$ are the last actions achieved (linear and angular velocities). The elementary observation vector $o_t$ is composed of depth values from the embedded Intel RealSense D435 sensor. The depth output resolution is $640\times480$ pixels with a maximum frame rate of $60$~fps. Since the depth field of view of this sensor is limited to $\ang{87}\pm{\ang{3}}\times\ang{58}\pm{\ang{1}}\times\ang{95}\pm{\ang{3}}$, the environment is only partially observable. To limit the amount of values kept from this sensor, we decided to sample $10$ values from a specific row of the depth map (denoted as $\delta$). 
By doing this, we sample the environment along the $(\vec{X};\vec{Y})$ orthonormal plane, similarly to a LIDAR sensor (but within an angle of~$\ang{58}$). In the following, the vector containing these $10$ depth values captured from a frame at timestep~$t$ is denoted by~$F_t$. To be able to avoid obstacles, it seems natural to consider a period of observation longer than the current frame. For this reason, three different observation vectors have been evaluated:
\begin{itemize}
\item The current frame only, $o_t^1=\left[F_t\right]$.
\item The three last frames, $o_t^2=\left[F_t \ ; \ F_{t-1}\ ; \ F_{t-2}\right]$
\item The last two frames and their difference,\\ $o_t^3=\left[F_t \ ; \ F_{t-1} \ ; \ (F_t-F_{t-1})\right]$
\end{itemize}
The \textit{sampling rate} for the training and prediction of a policy is a critical parameter. It refers to the average number of $s_t$ obtained in one second. If it is too high, the long-term effects of the actions on the agent state cannot be captured, whereas a too low value would most likely lead to a sub-optimal policy. A good practice is at the very least to synchronize the sampling process with the robot slowest sensor (by doing this, every state $s_t$ contains new information). This was the depth sensor in our case, which is also the main contributor to the observation vector.

\subsubsection{Policy structure}
\begin{figure}[b!]
\captionsetup{justification=justified}
\centering
\includegraphics[width=\linewidth]{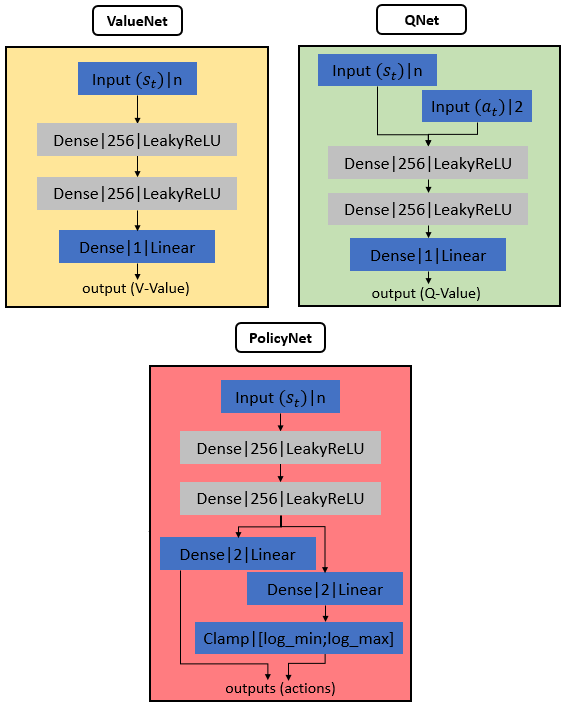}
\caption{The network structure for our implementation of the SAC algorithm. Each layer is represented by its type, output size and activation function. The dense layer represents a fully-connected neural network. The models use the same learning rate $l_r=3e^{-4}$, optimizer (Adam) \cite{Kingma2014AdamAM} and activation function (Leaky Relu, \cite{Maas2013RectifierNI}). The target smoothing coefficient $\tau$ is set to $5e^{-2}$ for the soft update and $1$ for the hard update.}
\label{fig1:sacNet}
\end{figure}

The architecture of networks encoding the policy seemed to have little impact, therefore we did not put a lot of emphasis on this part and a single one has been selected. In order to optimize the functions introduced in Section \ref{SAC}, three fully-connected neural networks are used as shown in Figure~\ref{fig1:sacNet}. The $n$-dimensional depth range findings, the relative target position and the last action achieved are merged together as a $(n+4)$-dimensional state vector~$s_t$. The sparse depth range findings are sampled from the raw depth findings and are normalized between~$0$ and~$1$. The 2-dimensional relative target position is represented in polar coordinates with respect to the robot coordinate frame. The last action performed takes the form of the last linear and angular velocities that are respectively expressed in $\mathrm{m.s}^{-1}$ and $\mathrm{rad.s}^{-1}$.

\subsubsection{Reward shaping}

Reinforcement learning algorithms are very sensitive to the reward function, which seems to be the most critical component before model transfer strategy. A straightforward sparse reward function (positive on success, negative on failure) would most likely lead to failure when working with a physical agent. On the other hand, a too specific reward function seems too hard to be learned. However, we describe below how the reward shaping approach~\cite{laud2004theory} could lead to an interesting success rate in simulation.

\noindent At the beginning of each episode, the robot is placed at an initial position~$P$ with a randomized orientation~$\theta$. The goal for the robot is to reach a target position $T$ whose coordinates ($x_T,y_T$) change at each episode (the target relative position from the robot is an input of the models). 
Reaching the target (considered achieved when the robot is below some distance threshold $d_{min}$ from the target) produces a positive reward $r_{reached}$, while touching an element of the environment is considered as failing and for this reason produces a negative reward $r_{collision}$. The episode is stopped if one of these events occurs. Otherwise, the reward is based on the difference $dR_t$ between $d_t$ (the Euclidean distance from the target at timestep $t$) and $d_{t-1}$. If $dR_t$ is positive, the reward is equal to this quantity multiplied by a hyper-parameter $C$ and reduced by a velocity factor $V_{r}$ (function of the current velocity $v_t$ and $d_t$). On the other hand, if $dR_t$ is negative (which means the robot moved away from the target during the last time step), the instant reward is equal to $r_{recede}$. The corresponding reward function is thus:
\begin{equation}
r(s_t,a_t)=
\left\{
\begin{aligned}
&C\times dR_t\times V_r\hspace{0.05in}\textbf{if}\hspace{0.05in}dR_t>0\\
&\hspace{10px}r_{recede}\hspace{20px}\textbf{if}\hspace{0.05in}dR_t\leq0\\
&\hspace{10px}r_{reached}\hspace{16px}\textbf{if}\hspace{0.05in}d_t<d_{min}\\
&\hspace{10px}r_{collision}\hspace{14px}\textbf{if}\text{ collision detected}
\end{aligned}
\right.
\end{equation}

\noindent where 
$ V_r=(1-\max(v_t,0.1))^{1/{\max(d_t,0.1)}}$,
${r_{reached}=500}$, $r_{collision}=-550$ and $r_{recede} = -10$.
\vspace{0.2cm}

\noindent Without this velocity reduction factor $V_r$, we observed during training that the agent was heading toward the target even though an object was in its field of view (which led to a collision). The reward signal based only on the distance rate $dR_t$ was too strong compared to the collision signal. 
With this proposed reward function, we encourage the robot to get closer to the target and to decrease its velocity while it gets to the goal. In addition, it is important to relate the non-terminal reward to the distance of the current state to the target. This way, it is linked to a state function (a potential function) which is known to keep the optimal policy unchanged. More precisely, if the reward was simply defined as $\gamma d_{t+1} - d_t$, then the optimal policy would be the same with or without the shaping (which just fastens the convergence). Here, the shaping is a little more complicated and may change the optimal policy but it is still based on $d_t$ (see \cite{Badnava2019ANP} for more details on reward shaping and its benefits).

\subsection{Incremental complexity vs naive Sim-to-Real transfer} \label{sec:training-setup}

\begin{figure}[!ht]
\captionsetup{justification=justified}
\centering
\includegraphics[width=\linewidth]{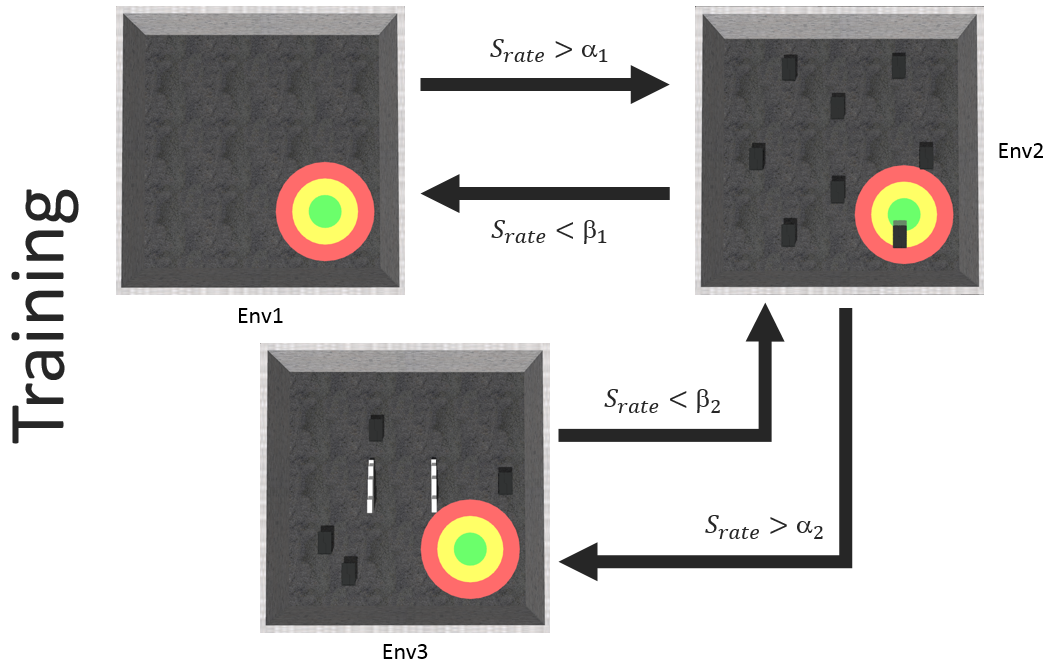}
\caption{Illustration of the incremental complexity strategy. The policy is trained on multiple environments, each one representing an increment of subtasks (more complex obstacles) contributing to the global mission.}
\label{fig1:Strategy}
\end{figure}

The mission was divided into three distinct environments as shown in Figure~\ref{fig1:Strategy}. The first one (\textit{Env1}) is an empty room. By starting training in this context, we try to force the agent to learn how to simply move toward the target. The second environment (\textit{Env2}) incorporates eight identical static obstacles uniformly spread in the room. Training in these conditions should allow the agent to learn how to avoid obstacles on its way to the target. The last environment (\textit{Env3}) includes both static and mobile obstacles. Two identical large static obstacles are placed near the initial position of the robot while four other identical mobile obstacles are randomly distributed in the room at the beginning of each episode. Transition from an environment to another is based on the success rate $S_{rate}$ for the last $100$ episodes. If this value exceeds a specific threshold, the agent will move to the next environment or will be sent back to the previous one. Transition from one environment to another is related to the local performance of the policy and is done during the current training session, ensuring the use of samples collected from various conditions to improve generalization.
As illustrated in Figure~\ref{fig1:Strategy}, $\alpha_1$ and $\beta_1$ rule transitions between \textit{Env1} and \textit{Env2} while $\alpha_2$ and $\beta_2$ rule transition between \textit{Env2} and \textit{Env3}. For this study, these parameters were set to $\alpha_1=90\%$, $\alpha_2=80\%$, and $\beta_1=\beta_2=50\%$. In the following, the ``naive" strategy refers to training using only either \textit{Env2} or \textit{Env3}. The training of all the models consisted of 5000 episodes with a maximum step size of 500 each. It was observed that learning with incremental complexity does not improve performance in simulation but has a critical impact in real life. It is relevant, since this domain randomization technique can be easily implemented for many other problems.

\section{\uppercase{Experiments}} \label{sec:expe_result}
\noindent Training or evaluating a robotic agent interacting with a real environment is not straightforward. Indeed, both the training (or at least the fine-tuning) and the evaluation require a lot of task runs. So in this work, we used both simulation and real-world experiments and particularly studied the behavior of the transferred policy from the former to the latter. To do so, a simulation environment representative of the real-world conditions was built, and the real world environment was also instrumented to carry out unsupervised intensive experiments. 

\subsection{Simulation experiments}

The proposed approach has been implemented using the Robot Operating System (ROS) middleware and the Gazebo simulator~\cite{Koenig2004DesignAU}. Some previous works already used Gazebo for reinforcement learning like Gym-Gazebo \cite{zamora2016extending,kumar2019offworld}. 
An URDF model representative of the true robot dynamics has been generated. The R200 sensor model from the Intel RealSense ROS package was used to emulate the depth sensor (at 10 fps), and the Gazebo collision bumper plugin served to detect collisions. We created several environments that shared a common base, a room containing multiple obstacles (some fixed, others with their positions randomised at each episode). The training process was implemented with \textit{Pytorch} \cite{NEURIPS2019_9015} as a ROS node communicating with the Gazebo node using topics and services.
Both the simulator and the training code ran on the same desktop computer equipped with an Intel Xeon E5-1620 (4C-8T, 3.5Ghz), 16GB of memory and a GPU Nvidia GTX 1080, allowing us to perform the training of one model in approximately $6$ hours in the Cuda framework \cite{Ghorpade2012GPGPUPI}. The communication between the learning agent and the environment was done using a set of ROS topics and services, which facilitated transposition to the real robot.

\subsection{Real-world experiments} \label{real:setup}

The real world experiment took place into a closed room measuring $7$ by $7$ meters. The room was equipped with a motion capture system (Optitrack) used by the robot and by a supervision stack (described in what follows). Four obstacles (boxes) were placed into the room at the front or the side of the robot starting point. The same desktop computer processed the supervisor and the agent. The robot used was a Wifibot Lab V4 robotic platform which communicated with the ground station using WiFi. It carried an Intel RealSense D435 depth sensor and an on-board computer (Intel NUC 7), on which the prediction was computed using the learned policy. 
Since the number of runs needed for training and validation is large, this raises some practical issues:
\begin{itemize}
    \item A long operation time is not possible with usual mobile robots due to their battery autonomy.
    \item Different risks of damaging the robotic platform can occur on its way to the target with obstacle avoidance.
\end{itemize}

\begin{figure}[ht!]
\captionsetup{justification=justified}
\centering
\includegraphics[width=\linewidth]{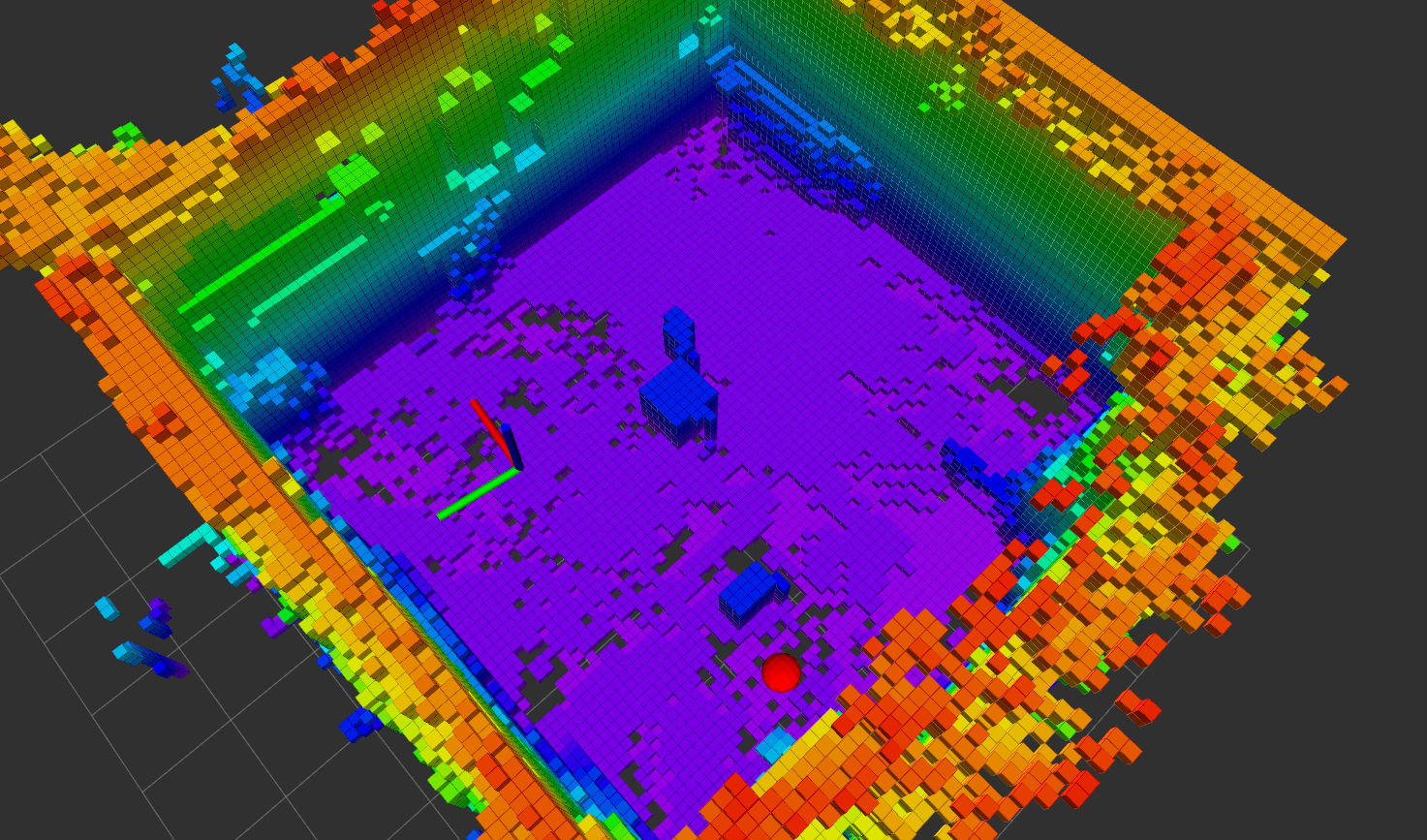}
\caption{Octomap ground truth of the environment. The frame denotes the robot position and the red ball the target position.}
\label{fig:octomap}
\hspace*{10px}
\end{figure}

\begin{figure}[t]
\captionsetup{justification=justified}
\vspace{0.25cm}
\includegraphics[width=\linewidth]{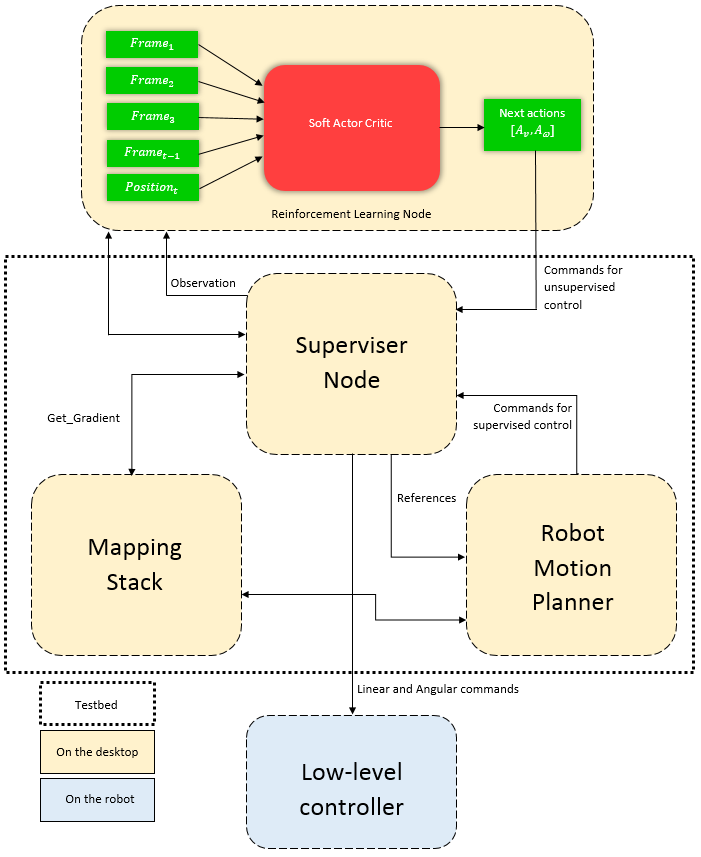}
\caption{Supervision stack for learning and testing in the real world.}
\label{fig1:superviser}
\end{figure}

\noindent To tackle these issues, we instrumented the environment with two components. First, the room setup allowed the robot to be constantly plugged into a power outlet without disturbing its movements. Secondly, we developed a supervisor node to detect collisions, stop the current episode, and replace autonomously the robot to its starting location at the beginning of a new episode. As detailed in Figure~\ref{fig1:superviser}, the supervisor multiplexes the command to the robot embedded low-level controller (angular and linear speeds). During the learning phase, it uses the command coming from the SAC node and during the resetting phase it uses the command coming from a motion planner node. The motion planner node defines a safe return trajectory using a PRM* path planner~\cite{sucan2012the-open-motion-planning-library} and a trajectory tracking controller~\cite{bak2001path}. No data was collected for learning during this return phase. During the episode, the supervisor node (Figure~\ref{fig1:superviser}) takes as input the linear and angular velocities estimated by our SAC model to send them to the robot. Whenever the episode is stopped, the supervisor takes as input the commands generated by the motion planner based on the mapping stack to make the robot move to a new starting position, without colliding with any element of the environment. Since the map is fixed, we built a ground truth 3D map (Figure~\ref{fig:octomap}) of the test environment before starting the experiment by manually moving the robot and integrating the depth sensor into an Octomap~\cite{wurm2010octomap} model (any other ground truth mapping technique would be suitable). Thanks to this infrastructure, we were able to run a large number  of  run-times  with a minimal need for human supervision. Obviously, the  duration  of  each  real-life  run  is  large  (vs simulation), but the unsupervised evaluation of 100 runs can be performed in roughly $\sim30$ minutes, which is practical for evaluating the Sim-to-Real policy transfer.

\subsection{Results} \label{results}
Experimental results for the approach proposed in the previous section are provided for the different observation vector configurations considered\footnote{\href{https://tinyurl.com/sim2real-drl-robotnav}{\scriptsize{A video can be found at https://tinyurl.com/sim2real-drl-robotnav}}}. This evaluation consisted in a total of 5 sessions of 100 episodes each, conducted with the real robot thanks to the supervision stack described in Section~\ref{real:setup}.
Let us stress that these performances are conservative due to safety margins included in the supervision stack but comparable for all models. It took us roughly $45$ minutes to test one model under these conditions. Performances of the trained policies were finally assessed and compared in terms of mean success rate and mean reward over the 5 sessions. These results are provided in Tables~\ref{results:sr} and~\ref{results:mr} for the distinct cases outlined in Section~\ref{sec:RLset}. In these tables, we designate by $F_n$ the $10$ depth values kept in the frame captured at time step $n$. This means that the first column indicates which observation vector $o_t^i$ is used in the state $s_t$. The second column specifies which environment has been used to train the models as shown in Section \ref{sec:training-setup}. It can observed that the models trained by using the incremental method (i.e. \textit{Env1-2-3} in the tables) obtain the best performances in terms of mean success rate as well as in mean reward over the 5 sessions. The best one among the models trained incrementally is the model whose observation vector consisted of the last two frames and their difference ($o_t^3$) with a success rate of $47$\% and a mean reward of $38.751$. The performance can thus be scaled twice using the incremental complexity sim-to-real strategy coupled with the SAC reinforcement learning strategy. This result is not trivial as depth-based mapless navigation is harder than mapless patrolling~\cite{zamora2016extending} or visual object recovery~\cite{sampedro2019fully}, which do not need to go close to obstacles (and these methods were only tested in simulated environments). It could be noted that even the naive learned-in-sim policy achieves a non-trivial success rate. The success rate could most probably be improved by carrying out a fine-tuning training session in the real-world experiment, however this is beyond the scope of this paper.

\renewcommand\theadalign{bc}
\definecolor{arylideyellow}{rgb}{0.91, 0.84, 0.42}

\begin{table}[h]
\centering
\vspace{0.25cm}
\caption{Success rate (in \%).\label{results:sr}}
{\small{$F_t$ designates depth measurements taken at time $t$.}}
{\renewcommand{\arraystretch}{1.25}
\begin{tabular}{|P{2.87cm}||P{1.75cm}|P{1.5cm}|}
  \hline
  \thead{Observation\\vector ($o_t$)} & \thead{Training\\environments} & \thead{Success\\Rate}\\
  \hline \hline
  $[F_t]$ & Env2 & 21\%\\
  $[F_t]$ & Env3 & 29\%\\
  \rowcolor{arylideyellow}
  $[F_t]$ & Env1-2-3 & 32\%\\
  $[F_t$ ; $F_{t-1}$ ; $F_{t-2}]$ & Env2 & 38\%\\
  $[F_t$ ; $F_{t-1}$ ; $F_{t-2}]$ & Env3 & 17\%\\
  \rowcolor{arylideyellow}
  $[F_t$ ; $F_{t-1}$ ; $F_{t-2}]$ & Env1-2-3 & 42\%\\
  $[F_t$ ; $F_{t-1}$ ; $F_{t}-F_{t-1}]$ & Env2 & 24\%\\
  $[F_t$ ; $F_{t-1}$ ; $F_{t}-F_{t-1}]$ & Env3 & 33\%\\
  \rowcolor{arylideyellow}
  $[F_t$ ; $F_{t-1}$ ; $F_{t}-F_{t-1}]$ & Env1-2-3 & 47\%\\
  \hline
\end{tabular}}
\end{table}
\begin{table}[h]
\caption{Mean reward values.}\label{results:mr} \centering
{\small{$F_t$ designates depth measurements taken at time $t$.}}
{\renewcommand{\arraystretch}{1.25}
\begin{tabular}{|P{2.87cm}||P{1.75cm}|P{1.5cm}|}
  \hline
  \thead{Observation\\vector ($o_t$)} & \thead{Training\\environments} & \thead{Mean\\reward}\\
  \hline \hline
  $[F_t]$ & Env2 & -248.892\\
  $[F_t]$ & Env3 & -189.68\\
  \rowcolor{arylideyellow}
  $[F_t]$ & Env1-2-3 & -95.623\\
  $[F_t$ ; $F_{t-1}$ ; $F_{t-2}]$ & Env2 & -100.662\\
  $[F_t$ ; $F_{t-1}$ ; $F_{t-2}]$ & Env3 & -300.124\\
  \rowcolor{arylideyellow}
  $[F_t$ ; $F_{t-1}$ ; $F_{t-2}]$ & Env1-2-3 & 22.412\\
  $[F_t$ ; $F_{t-1}$ ; $F_{t}-F_{t-1}]$ & Env2 & -217.843\\
  $[F_t$ ; $F_{t-1}$ ; $F_{t}-F_{t-1}]$ & Env3 & -56.288\\
  \rowcolor{arylideyellow}
  $[F_t$ ; $F_{t-1}$ ; $F_{t}-F_{t-1}]$ & Env1-2-3 & 38.751\\
  \hline
\end{tabular}}
\end{table}

\section{\uppercase{Conclusions}}
\noindent In this paper, we have proposed a mapless navigation planner trained end-to-end with Deep Reinforcement Learning. A domain randomization method was applied in order to increase the generalization capacities of the policy without additional training or fine-tuning in the real world. By taking as inputs only two successive frames of 10 depth values and the target position relative to the mobile robot coordinate frame combined with a new incremental complexity training method, the given policy is able to accomplish depth-based navigation with a mobile robot in the real world even though it has only been trained in a ROS-Gazebo simulator. When compared to a naive training setup, this approach proved to be more robust to the transfer on the real platform. The models trained in this study were able to achieve the mission in an open environment containing box-size obstacles and should be able to perform well in similar indoor contexts with obstacles of different shapes. However, they would most likely fail in environments such as labyrinths because the observation inputs $o_t$ will be too different. A direct improvement could be to include a final reference heading which can be easily considered since the Wifibot Lab V4 robotic platform is able to spin around. Future work will focus on the fair comparison between model-based methods and such learning algorithms for autonomous robot navigation, as well as addressing more complex robotics tasks.
\vspace{-0.01cm}
\bibliographystyle{apalike}
{\small\bibliography{ref}}

\end{document}